\documentclass[runningheads]{llncs}
\usepackage{graphicx}
\usepackage{amsmath,amssymb} % define this before the line numbering.
\usepackage{color}\usepackage[width=122mm,left=12mm,paperwidth=146mm,height=193mm,top=12mm,paperheight=217mm]{geometry}

\begin{document}
% \linenumbers
\pagestyle{headings}
\mainmatter

\title{Computational Beauty: Aesthetic Judgment at the Intersection of Art and Science} % Replace with your title

\titlerunning{Computational Beauty: Aesthetic Judgment at ...}

\authorrunning{E. L. Spratt and A. Elgammal}

\author{Emily L. Spratt$^\dag$ and Ahmed Elgammal$^\ddag$ }
\institute{$^\dag$ Dept. of Art and Archaeology, Princeton University, NJ, USA \\
$^\ddag$ Dept. of Computer Science, Rutgers University, NJ, USA}

\maketitle

\begin{abstract}
In part one of the {\em Critique of Judgment}, Immanuel Kant wrote that ``{the judgment of taste~\ldots~is not a cognitive judgment, and so not logical, but is aesthetic}~\cite{Kant}.'' While the condition of aesthetic discernment has long been the subject of philosophical discourse, the role of the arbiters of that judgment has more often been assumed than questioned. The art historian, critic, connoisseur, and curator have long held the esteemed position of the aesthetic judge, their training, instinct, and eye part of the inimitable subjective processes that Kant described as occurring upon artistic evaluation. Although the concept of intangible knowledge in regard to aesthetic theory has been much explored, little discussion has arisen in response to the development of new types of artificial intelligence as a challenge to the seemingly ineffable abilities of the human observer. This paper examines the developments in the field of computer vision analysis of paintings from canonical movements within the history of Western art and the reaction of art historians to the application of this technology in the field. Through an investigation of the ethical consequences of this innovative technology, the unquestioned authority of the art expert is challenged and the subjective nature of aesthetic judgment is brought to philosophical scrutiny once again.

\end{abstract}

\section{Aesthetics: Between Computer Science and Art History}

%Since James Gibson's pioneering research on two-dimensional imaging for statistical pattern recognition, which brought the computer from a typewriting calculator to an image-processing machine, computer vision science has developed into an independent field of study within the quickly evolving domain of artificial intelligence. 

Since the pioneering research on two-dimensional imaging for
statistical pattern recognition that took place in the 1960s, when the
computer was brought from a typewriting calculator to an
image-processing machine, the field of computer vision science has
developed into an independent field of study within the quickly
evolving domain of artificial intelligence.
While developments within computer vision have mainly derived from the impetus of defense technology, in the last twenty years the application of this research has been applied to the interpretation of two-dimensional images, creating a new branch of study. For example, computer vision utilizes algorithms for different object recognition related problems including: instance recognition, categorization,  scene recognition, and pose estimation. At this point in time, computers can examine an image and recognize distinct objects, and even categorize the scenes they occupy. Cultural and historical inferences about an image may slowly become determinable by computers, yet the complexities of these higher-level perceptions are currently possible only in the realm of human cognition.

On account of the significant advances in computer vision research in the analysis of art, we would like to suggest that the time has come to make an overall evaluation of the possibilities of aesthetic interpretation that the computer offers to date. While academics in the humanities have remained largely skeptical of the use of computer science to perform tasks that involve subjective interpretations of qualitative data, we seek to demonstrate how one intersection of the arts and sciences can be fruitfully navigated, that of computer vision and art history. Rather than relegating the aesthetic interpretation of art by computers solely to computer scientists, let us determine how machine-based analysis of art functions in comparison to human judgment by considering the voices of art historians and other representatives from the humanities. This collaborative approach thus heralds a reevaluation of the philosophy of aesthetic theory as it has been applied in art history in light of the scientific developments not only within computer vision but also in relation to neurobiology.\footnote{See, for instance, New York University's Center for Neural Science and the Visual Neuroscience Laboratory, \url{http://www.cns.nyu.edu/}.}  

Indeed, computer vision challenges the art historian's very conception of the processes of aesthetic judgment and what may be regarded as objective or subjective mental processes if a computer has the ability to perform similar tasks. Through examination of the innovations and histories of computer vision and aesthetics as a philosophical discourse that has been utilized in art history, we will question both how notions of authority in aesthetic judgment and the processes of aesthetic interpretation itself have been and are being constructed. While the art historian, critic, connoisseur, and curator have long held the esteemed position of aesthetic judge--their training, instinct, and eye, part of a seemingly inimitable cognitive process that occurs upon artistic evaluation--these new developments in computer science challenge the very tenets of aesthetic theory and call for their reevaluation. Similarly, this paper demands an accessible explanation from computer scientists as to how aesthetic judgments are being programmed into machines and to what end. Through a collaborative approach, we aim to begin to bridge the gap between computer science and art history, fostering research that will yield effective applications of computer vision in the analysis of art and theoretical reconsideration of aesthetic judgment given the newfound capabilities of machines.

In this paper, we will question the potential of a computer to make aesthetic judgments. We will consider the degree to which computers can aid specialists within art history and examine whether computer vision can offer unique insights to art historians regarding iconographic and stylistic influence. We also will examine whether art historians would be open to using new technologies advanced by these developments in computer science and offer suggestions as to how to encourage collaboration between the fields. Through the initiation of a multidisciplinary discussion about these interrogations, this approach to two seemingly disparate fields, to our knowledge, is the first of its kind. The paper's structure is as follows: Sections two and three will review the research developments in computer vision regarding the analysis of art and examine the reaction of art historians to these developments. We explain the philosophical concept of aesthetic judgment and its implications in sections four and five. In the conclusion, we will
discuss the present and future interaction between the fields of
computer science and art history.

\section{Computer-based Stylistic Analysis of  Art}

%TBA
%\section{Computer-based Analysis of Art: A Review of Recent Innovations in the Field}

The field of computer vision is focused on developing algorithms for understanding images and videos using computers and providing interpretations of them, essentially giving computers the ability to see. Given the context of the application, these interpretations have the capacity to yield highly variegated meanings, including the ability to recover three-dimensional forms of representations from a two-dimensional image, the recognition of objects in an image, and the analysis of human activities, gestures, facial expressions, and interactions.

In the last two decades, within the field of computer vision, there has been increasing interest in the area of computer-based analysis of art with some degree of collaboration with art historians. Earlier work in this area has focused on providing objective analysis tools, where computers are mainly used to quantify certain physical features of an artwork. These tools can provide art historians with measurements that are difficult to obtain by the human eye alone. For example, computer vision technology has been used to conduct extremely precise pigmentation analysis of a painting's color, quantify exacting statistical measures of brushstrokes, and provide detailed examinations of craquelure~\cite{stork2009computer}. Computers also can provide tools to automate certain types of analysis that have long been performed manually by experts, particularly in the interpretation of perspective and lighting, and the decipherment of anamorphic images. An approachable review of the research in this area prior to 2009 was conducted by David Stork~\cite{stork2009computer}.

With the advances in computer vision and machine learning, computers now can make semantic-level predictions from images. For example, computers can now recognize object categories, human body postures, and activities in a scene. As a result, research on computer-based analysis of art has evolved and is now developing more sophisticated tasks, including the automatic classification of art to identify the hand of an artist, the ability to classify paintings according to style and to distinguish stylistically similar images of paintings, the quantification of the degree of artistic similarity found between paintings, and the capability to predict a painting's date of production. We collectively call these tasks computer-based stylistic analysis of art. At this point in time, computer vision has gone far beyond providing art historians with tools that are simply stylometric, or quantifiable physical measures. One trend in computer vision technology is the development of algorithms that encompass complex measures taken through a computer's visual analysis that are used to directly make predictions about a painting's attribution, date, authenticity, and style without the need of an art historian. In this section we review some of these new developments within computer science that approach the realm of aesthetic judgment through computer automation.
\\
\indent Most of the research concerning the classification of paintings utilizes low-level features or simple diagnostic measures, such as the appearance of color, shadow, texture, and edges. Researchers have extensively conducted computerized analysis of brushstrokes in images of paintings~\cite{sablatnig,li2004studying,lyu2004digital,johnson2008image,berezhnoy2009automatic,brdahujapo09,Jia12}. Brushstrokes, like fingerprints, provide what computer scientists call a signature that can help distinguish the hand of the artist. The analysis typically involves texture features that are assumed to encode the brushstroke signature of the artist. Recently, Li et al. proposed a method based on the integration of edge detection and image segmentation for brushstroke analysis~\cite{Jia12}. Using these features they found that regularly shaped brushstrokes are tightly arranged, creating a repetitive and patterned impression that can represent, for example, Van Gogh's distinctive painting style, and help to distinguish his work from that of his contemporaries. This research group has analyzed forty-five digitized oil paintings of Van Gogh from museum collections. 
\\
\indent T.E. Lombardi has presented a study of the capability of different types of low-level features extracted from paintings to identify artists~\cite{Lombardi}. Several features such as color, line, and texture were surveyed for their accuracy in classification of a given painting to identify the hand of an artist amongst a small data set of artists. Additionally, several machine learning techniques were used for classification, visualization, and evaluation. Through this research, the style of the painting was identified as a result of the computer's ability to recognize artistic authorship. For example, recognition that a painting was attributed to Claude Monet signaled an association with Impressionism. The idea of using color analysis for the identification of a painter has also been researched~\cite{widjaja}. Bag of Words (BoW) (an approach originally used a decade ago for text classification and object recognition) was utilized by Khan et al. along with the fusion of color and shape information that could identify individual painters~\cite{khan}. 
\\
\indent  The problem of annotating digital images of art prints (painted copies of canonical paintings) was addressed by Carneiro et al.\cite{Carneiro11}. In that research, a reproduction was automatically annotated to one of seven themes (e.g., the Annunciation) as well as with the appearance of twenty-one specific symbols or objects (e.g., an angel, Christ, Mary). These computer scientists proposed that a graph-based learning algorithm, based on the assumption that visually similar paintings share the same types of annotation, would yield higher levels of accuracy in the identification of paintings. The data set they used contained reproductions from the fifteenth to the seventeenth century that were annotated by art historians and focused exclusively on religious themes. The analysis of art print images was later extended using a larger data set (PRINTART) with semantic annotation (e.g., Holy Family), localized object annotation, such as localizing a rectangle around the Christ Child, and simple body pose annotation, for instance locating the head and torso of Mary~\cite{Carneiro12}. The research of Carneiro et al. demonstrated that the low-level texture and color features, typically exploited for photographic image analysis, are not effective because of inconsistent color and texture patterns describing the visual classes in artistic images~\cite{Carneiro12}. In essence, the quality of painted reproductions greatly affects the ability of a computer to visually interpret a painting.
\\
\indent The research of Graham et al. examined the way we perceive two paintings as similar to each other~\cite{Graham10}. The researchers collected painting similarity ratings from human observers and used statistical methods to find the factors most correlated with human ratings. They analyzed two sets of images, denoted as either scenes of landscapes or portraits and still lives. The analysis demonstrated that similarities between paintings could be interpreted in terms of basic image statistics. For landscape paintings, the image intensity statistics were shown to highly correlate with the similarity ratings; for portraits and still lives, the most important visual clues about their degree of similarity were determined to be semantic variables, such as the representation of people in a given composition.
\\
\indent The question of automatically ordering paintings according to their date of production was posed by Cabral et al.\cite{CabralCDBC11}. They formulated this problem by embedding paintings into a one-dimensional linear ordering and utilizing two different methods. In the first, they applied an unsupervised (without the use of annotation) dimensionality reduction (a technique used in machine learning to reduce the number of variables). To do so, they only needed to employ visual features to map paintings to points on a line. This approach, despite being fast and requiring no annotation, resulted in low accuracy. The second method took into account available partial ordering of paintings annotated by experts. This information was used as a constraint in order to find the proper embedding of a painting to a line, which was more chronologically accurate. 
\\
\indent 
Unlike most of the previous research that focused on inferring the authorship of the artist from the painting, Arora et al. approached the problem of the classification of style in paintings into classes that are directly recognized in the history of art~\cite{Arora12}. They defined a classification task between seven painting styles: Renaissance, Baroque, Impressionism, Cubism, Abstract, Expressionism, and Pop Art. In their research, they formulated a supervised classification problem (a machine learning paradigm where training data is assumed to have class labels annotated by experts). They presented a comparative study evaluating generative models versus discriminative models, as well as low- and intermediate-level versus semantic-level features. For the semantic-level description they used features called Classeme, which encode an image in terms of the output of a large number of classifiers~\cite{Torresani2010}. Such classifiers are trained using images retrieved from Internet search engines, with an accompanying term list. The result was particularly interesting: the research found that the semantic-level discriminative model produced the best classification result with 65\% style classification accuracy~\cite{Arora12}. Indeed, the use of verbal descriptors that are associated with the visual content of a painting led to greater accuracy in classification compared to stylistic analysis alone. This result highlights the importance of encoding semantic information for the task of style classification and for the analysis of art in general. 
\begin{figure}[t]
\centering
\includegraphics[width=0.75\linewidth]{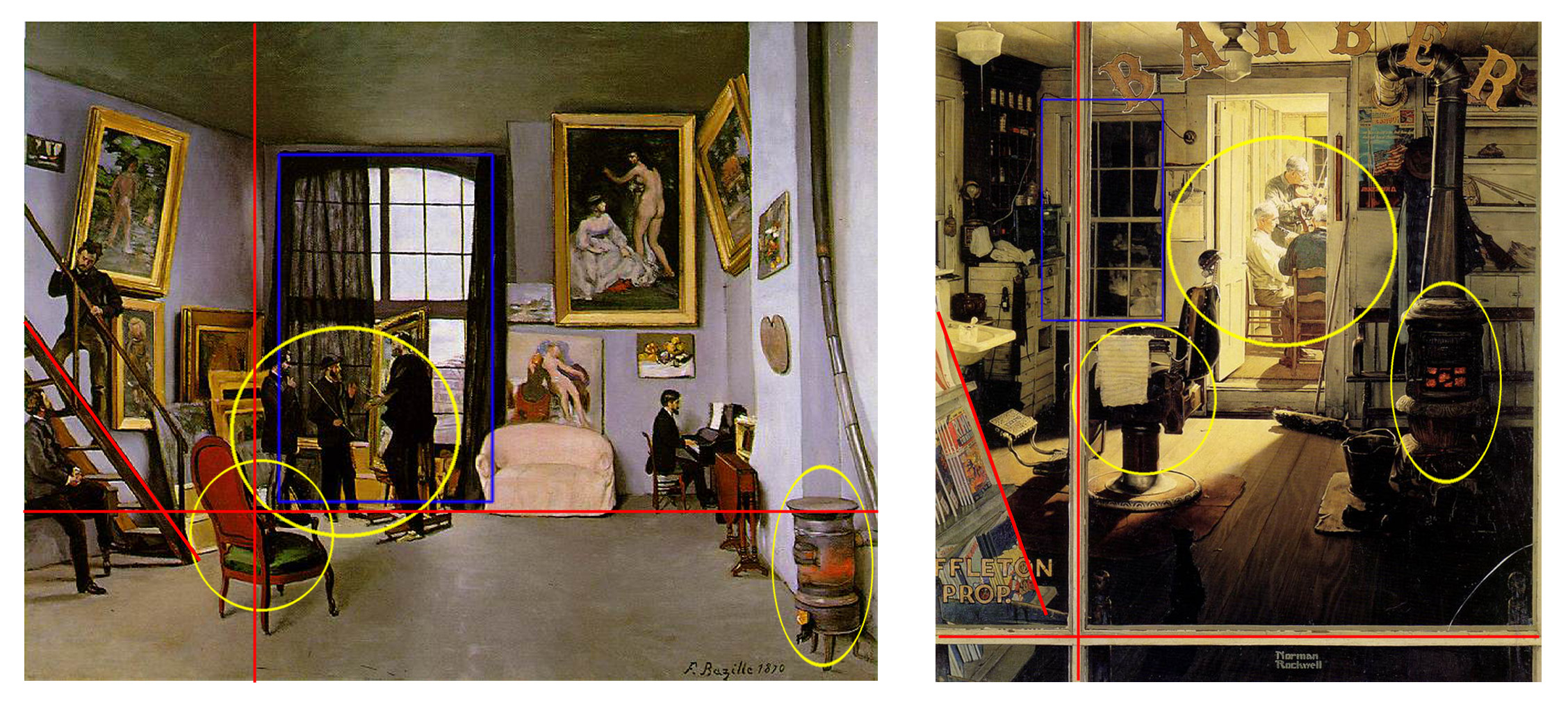}
\caption{A computer-recognized example of stylistic similarity, from Abe et al. is 
Fr\'ed\'eric Bazille's \textit{Studio 9 Rue de la Condamine} (left), Norman Rockwell's \textit{Shuffleton's Barber Shop} (right)~\cite{Abe2013MM4CH}. }
%The composition of both paintings is divided in a similar way. Yellow circles indicate similar objects, red lines indicate compositional analysis, and the blue square represents similar structural elements. The recognizable objects include a fire stove, three men clustered, chairs, and a window. These features are seen in both paintings and in similar locations in the compositions. Although there is no direct connection between Bazille and Rockwell in the history of art, the computer was able to detect the purely stylistic similarities of the paintings. How these shared formal characteristics, which occurred by coincidence, can be useful to an art historian remains unclear. This type of pairing, however, may be useful to studio artists interested in examining different types of representational strategies across time and space.}
\label{fig:figure2}
\end{figure}
\indent The problem of discovering similarities between artists and inferring artistic influences was addressed by Abe et al. by defining similarity measures between artists over a data set of sixty-six artists and 1,710 paintings, ranging from the fifteenth to the twentieth century~\cite{Abe2013MM4CH,Saleh2014}. Based on the results of the research of Arora et al., they also used semantic-level features to encode the similarity of paintings~\cite{Arora12}. Artist-to-artist similarity was encoded with variants of the Hausdorff distance (a regularly used geometric distance measure between two sets of points). This similarity measure was utilized to construct a directed graph of artists encoding both artist-to-artist similarity and temporal constraints, and that graph was used to discover potential influences. They evaluated their results by comparing the discovered potential influences against known influences cited in art historical sources. Figure~\ref{fig:figure2} illustrates an example of two stylistically similar paintings detected by the approach of Abe et al., Fr\'ed\'eric Bazille's \textit{Studio 9 Rue de la Condamine} (1870) and Norman Rockwell's \textit{Shuffleton's Barber Shop} (1950)~\cite{Abe2013MM4CH,Saleh2014}. This type of comparison, however, would not be cited in art historical sources, as the connection between the paintings is purely formal and coincidental. The graph of artists was also used to achieve a visualization of artistic similarity (this is termed, map of artists). 
\\
\indent Most of the aforementioned research uses computer vision analysis to perform tasks implicitly related to the domain of aesthetic judgment. There has also been recent research that has developed algorithms to make aesthetic judgments of a more explicit nature~\cite{datta2006studying,dhar2011high}. This research has used computer vision technology to predict how humans would score an image of a scene or an object according to its perceived beauty. For example, computer models can be trained to predict attributes of an image that beg aesthetic discernment, such as compositional strategies, the presence of particular objects, and even the way a sky is illuminated. The attributes are then used to predict aesthetic calculations for a given image~\cite{dhar2011high}. This type of computer vision analysis narrows the concept of aesthetic judgment to a set of pre-defined objective rules.\footnote{Ethical consideration of the use of computer vision technology for
these purposes is clearly needed and requires further investigation.}  
\\
\indent Given the progress in developing computer algorithms that are directly related to tasks regarding what humans would define as aesthetic judgments, a number of questions emerge regarding the implications of these applications. The ability of these computational models to perform aesthetic judgments in this capacity demonstrates that there is a difference in perception of the processes required for artistic evaluation in the arts versus the sciences. Currently, there is a trend in artificial intelligence and computer vision technology in using computational models inspired by the brain's complex neural network (known as deep networks); as the similarities between computer systems and neurobiology expand, the differences between aesthetic interpretation as it is understood in the humanities as opposed to the sciences will only widen if the questions this research poses are not adequately addressed. In the following sections we therefore explore the implications of these developments in computer vision technology.

\section{Perspectives from the Field of Art History: Does Computer Vision Pose a Threat?}

Unfortunately, these developments in computer vision are not widely known or fully understood in the humanities and thus indicate the disjuncture between the fields of art history and computer science, and a larger fracture between the arts and sciences. In order to gauge the current perceptions between these fields, we conducted two surveys that were distributed to computer scientists and art historians at Princeton University, Rutgers University, Cornell University, New York University, and the University of California at Los Angeles in August 2014.\footnote{{ \raggedright For a complete analysis of our digital humanities survey see}\\ \url{https://sites.google.com/site/digitalhumanitiessurvey/}}  The results revealed that while there has been some positive reception of the use of computer vision research in art history, it remains limited and often confined to the domain of art conservation and connoisseurship. Not only is there a general unfamiliarity with the developments of new technologies like those in computer vision and their potential use-value in the humanities, there is much concern about their implementation. While it is not surprising that the majority of computer scientists thought that the use of artificial intelligence technology in the humanities signals the beginning of a positive paradigm shift in academia whereas the majority of art historians thought it did not, we also discovered that computer scientists and art historians are, in fact, in agreement on key issues. For instance, both groups agreed that they should collaborate and that computer vision technology does not risk taking away an art historian's job.\footnote{Sensationalizing titles about computer vision research in the press may be inaccurate. See, Matthew Sparkes, ``Could computers put art historians out of work?''~\cite{Telegraph}.}  Similar observations regarding the anxieties about the digital humanities project have been noted in regard to other specific applications.\footnote{ These concerns are well expressed in, Stephen Marche, ``Literature is not Data, Against Digital Humanities''~\cite{Marche2012}.} Although art historians are generally skeptical of allowing computers to perform tasks that have been traditionally reserved for trained specialists and deemed capable for only human comprehension, to date there has been, to our knowledge, no exact measures of this implicit distrust in the sciences to produce knowledge of a subjective nature.\footnote{See, for instance, Stanley Fish, ``Mind Your P's and B's: The Digital Humanities and Interpretation''~\cite{Fish2012}.}
\\
\indent Indeed, the key question in our survey, which inquired whether art historians would be willing to use computer vision to better understand paintings, aroused a strong territorial response from the field of art history. Given our empirical measures, which clearly demonstrate the divide between the fields, the authors of this paper have been trying to build a bridge between the disciplines that addresses the specific misunderstandings the survey results revealed. By investigating and analyzing the consequences that the use of artificial intelligence in domains that are traditionally understood to be reserved for humans pose, we hope to prevent further sequestration between the fields of art history and computer science. 
\\
\indent What does it mean when an art historian, who is trained to evaluate art, or even a novice admirer of art, is faced with a machine that can perform a similar task? Since the very nature of our ability to aesthetically comprehend and judge beauty is the determining factor in what most people would describe as distinguishing us from machines, this type of computer science threatens our own conceptions of human identity~\cite{paul2014}.  While it is important to recognize these anxieties, we would like to propose that understanding some of the philosophical origins of how we have come to regard aesthetic judgment may offer a partial explanation as to why it is that persons not trained in computer science perceive these developments as a threat. Computer science, neither our friend, nor foe, presents to the humanities a challenge: is intangible, or sensory, knowledge really intangible if a computer can perform processes that manifest the same results that a human would produce?

\section{Aesthetic Judgment: Between Philosophy and Art History} 

The concept of sensory knowledge derives from a long tradition in European theology, philosophy, and psychology, although it was not until the eighteenth century that this type of knowing began to be perceived in a positive light~\cite{preziosi1998art}.  Predominantly on account of Alexander Gottlieb Baumgarten's {\em Aesthetica}, published in Latin in 1750, the notion that there was a type of knowledge distinct from that of logic or reason gained acceptance~\cite{Baumgarten}. He termed this knowledge as {\em analogon rationis}, or analogue of reason, which had its own perfection distinct from logic. In consequence to this theory, it came to be argued that there should be two kinds of corresponding sciences of knowledge: that of logic and that of aesthetics. Baumgarten's philosophy thus provided the foundation for Immanuel Kant's theories on aesthetics and the background for the {\em Critique of Judgment}, published in 1790~\cite{Kant,preziosi1998art}.  
\\
\indent The key to Kant's discourse was his rooting of the condition of aesthetic discernment in a subjective, non-logical process. Indeed, the philosophy of aesthetics from Baumgarten to Deleuze, not necessarily including the branch of philosophy that Hegel directed aesthetics, places aesthetic comprehension in the realm of subjectivity.\footnote{Hegel regarded art as ``a secondary or surface phenomenon... thus harking back to pre-Baumgarten and pre-Kantian ideology which privileged the ideal or Thought by devalorizing visual knowledge.'' See Preziosi~\cite{preziosi1998art}, {\em The Art of Art History}, 66-67.}
Kant articulated the conditions of this type of reasoning in the {\em Critique of Judgment}, locating aesthetic understanding in moral philosophy and the principles of universality~\cite{Kant}.  
In part one of the {\em Critique}, Kant explains the processes of analysis that is required for the interpretation of art. He writes:
\begin{quote}
{ If we wish to discern whether anything is beautiful or not, we do not refer the representation of it to the Object by means of understanding with a view to cognition, but by means of the imagination (acting perhaps in conjunction with understanding) we refer the representation to the Subject and its feeling of pleasure or displeasure. The judgment of taste, therefore, is not a cognitive judgment, and so not logical, but is aesthetic- which means that it is one whose determining ground cannot be other than subjective.
\footnote{Kant, {\em Critique of Judgment}~\cite{Kant}, 41.}}
\end{quote}

Despite the focus on the subjectivity of aesthetic interpretation through individual judgment, Kant goes on to explain that the judgment of taste is also universal. He considers this in regard to the knowledge of how things are, or their ``theoretical knowledge,'' and to how things should be, or their  ``morality.''\footnote{Ibid., this interpretation was facilitated by Donald Preziosi, see Preziosi [27], {\em The Art of Art History}, 66-67.}
Kant argues that judging art is like judging the purposiveness of nature, as both can be examined in terms of beauty, either natural or artistic. While the philosophical relationship of nature and art remain outside the confines of this paper, it is important to take note that art was often evaluated in terms of its faithfulness to imitating nature until the modernist revolution led to the questioning of these values.
\\
\indent Just as nature was judged in terms of its purposiveness and its ability to manifest this quality in visual form, so too was art through its references. In this sense, Kant's perception of the quality of art is bound to the principles of the Romantic movement, as art historian Donald Preziosi notes, ``the world being the Artifact of a divine Artificer~\cite{preziosi1998art}.'' Positioning himself against classical rationalism, that beauty is related to a singular inner truth in nature, Kant instead suggests that beauty is linked to the infinite quality of the human imagination yet grounded in the finiteness of being. In this sense, the universality of taste also relates to a type of collective consciousness that stems from God's universal creation. Kant further relates aesthetics and ethics, positing that beautiful objects inspire sensations like those produced in the mental state of moral judgment, thus genius and taste could be related to the moral character of an artist or viewer. How moral values can raise or lower the aesthetic value of art is, indeed, a subject of philosophical scrutiny, if not controversy, to this day~\cite{mcmahon2013art}.  
\\
\indent The direction that Kant steered aesthetics has had a pervasive influence in philosophy into the contemporary period as Gilles Deleuze's conception of a transcendental empiricism demonstrates in its use of Kantian notions of sensibility. While art history has a tradition of intellectual borrowings for its theories and methodologies, its montage nature as a discipline, incorporating the perspectives of diverse fields in the humanities such as philosophy, comparative literature, anthropology, archaeology, and psychology, to name a few, has allowed for its inherent flexibility in critical interpretations that rarely produce a singular analysis of art. Indeed, parallel interpretations of a given object are implicitly understood to exist stemming from a wide range of theories and methodologies such as formal analysis, studies in iconography, conservation history, connoisseurship, Marxist theory, feminist theory, or social history, to list just several art historical perspectives, all of which may overlap or exclude each other. 
 \\
\indent Although the birth of art history is usually associated with the Renaissance and Giorgio Vasari's writing of the {\em Lives of the Most Excellent Painters, Sculptors, and Architects}, first published in 1550, how we define the origins of the discipline differs greatly according to the artistic tradition being considered, thus nuancing any standardization of what is meant by art historical analysis. In the West, Greek philosophers such as Plato and Aristotle could be credited with engaging in an early form of art history, commenting at length on the faculties of observation gained through sight and the physical drives associated with seeing~\cite{Aristotle_Poetics}. Indeed, throughout the history of the discipline, art history has been directly influenced by the sciences to varying degrees over time and according to geography, yet never to the exclusion of philosophical approaches to the interpretation of art. For example, Carl Linnaeus (1707-1778), the founding father of modern taxonomy who drew heavily from Francis Bacon's (1561-1626) scientific method of empiricism, may be credited with establishing the foundations for the classification of artifacts in museums through his organization of natural history objects concurrently with philosophical developments in art history~\cite{Zittel2008,Findlen1994}. 
\\
\indent It wasn't until the nineteenth century, however, that the principles of connoisseurship that emerged from Vasari's legacy were reevaluated by Giovanni Morelli (1816-1891)~\cite{Fernie1995}. While the period from the sixteenth century to the end of the nineteenth century witnessed many methodological developments in the history of art, these contributions were largely philosophical and less emulative of the direction taken by Linnaeus. Morelli's innovation was to focus on methods of connoisseurship that privileged direct engagement with a work of art that allowed for a very precise type of visual investigation. For instance, the rendering of a detail such as an ear could reveal the true authorship of a painting~\cite{Morelli}. 
\\
\indent Morelli writes in a dialogue from {\em Italian Painters}, published in 1890, ``Art connoisseurs say of art historians that they write about what they do not understand; art historians, on their side, disparage the connoisseurs, and only look upon them as the drudges who collect materials for them, but who personally have not the slightest knowledge of the physiology of art~\cite{Morelli2}.''  Morelli, and later the Vienna School of art history, which was heralded by Alois Riegl's (1858-1905) contributions on the history of ornament in terms of form (as opposed to history or philosophy), emphasized the strictly material interpretation that art history also accommodates. Not surprisingly, the theories of art espoused by Morelli and Riegl found immediate application to the world of connoisseurs, conservators, and museum associates. In the same vein, these types of materialist inquiries opened theoretical ground for philosophical consideration of the history of art measured through the development of form itself, devoid of its socio-historical constraints.  
\\
\indent This brief review of some of the intersections between art history and the sciences, both in terms of the faculties of vision and aesthetic judgment along with the field's engagement with scientific methodologies, underscores the point that there has been a sustaining influence of science in the arts. Therefore, if we were better able to understand the capabilities of computer vision technology, why wouldn't art historians consider the philosophical implications of this modern-day science on aesthetic theory and visual perception?

\section{The Implications of Aesthetic Philosophy on Human Perception and Art History}

\begin{figure}[t]
\center
\includegraphics[width=0.4\linewidth]{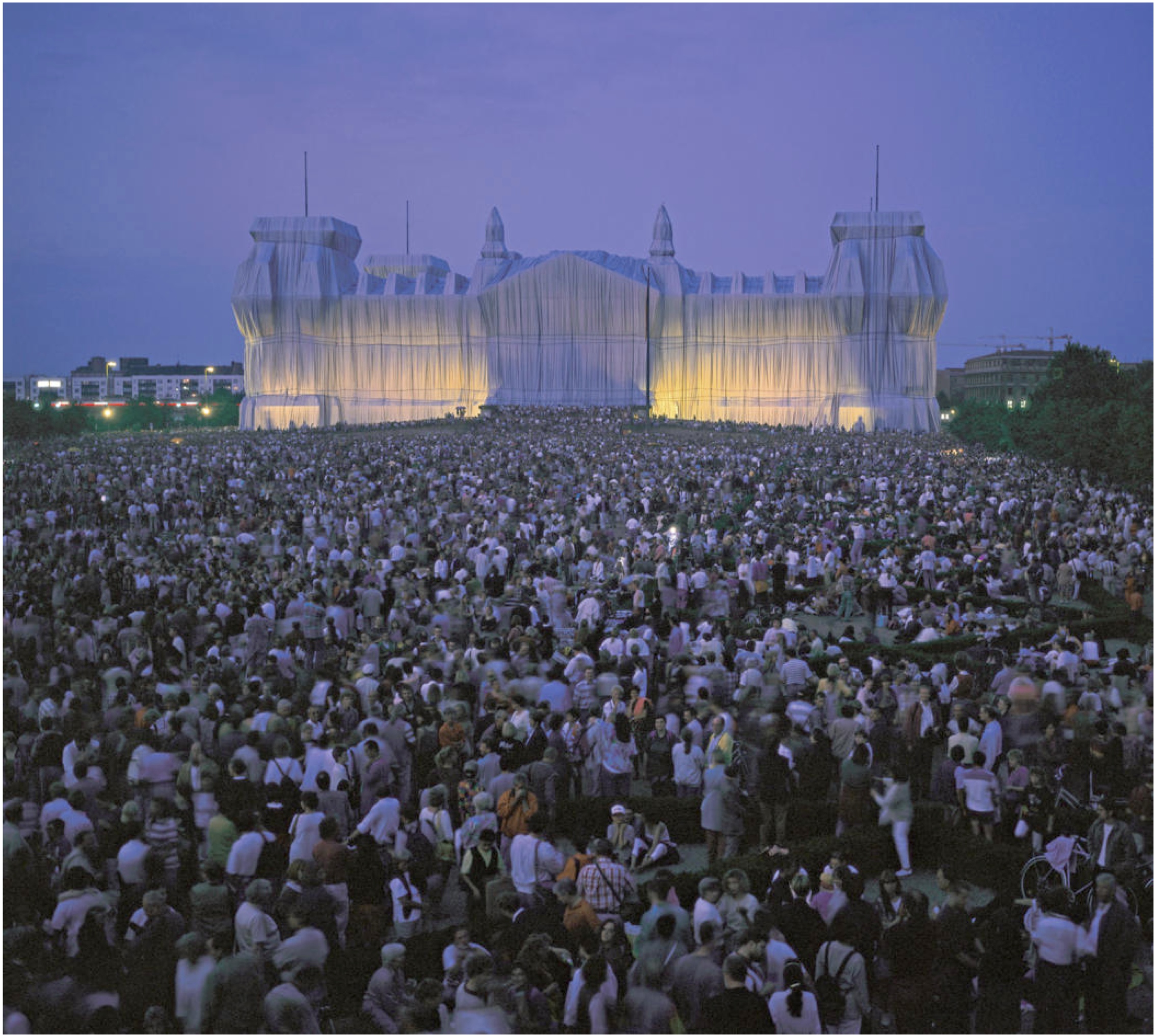}
\caption{{\em Wrapped Reichstag}, Berlin, 1971-95 Christo and Jeanne-Claude, Photo: Wolfgang Volz 
\copyright 1995 Christo}
\label{WrappedReichstag}
\end{figure}

The machine's ability to make an aesthetic judgment about a painting, and then compare it stylistically to other paintings, demonstrates that logic is at work in the complicated algorithms that comprise the artificial intellegence system. These processes are all clearly imitative and objective at the point of the computer program training period; once the machine reaches the automaton level, the question of subjectivity enters. In this sense, are computer programmers like blind watchmakers, to use Richard Dawkins' famous metaphor of the evolution of the universe and the free will debate~\cite{dawkins1996blind}? Are computers comparable to humans with genetic codes that predetermine outcomes, which are then shaped by the environment? 
\\
\indent While structural similarities between the human brain and computer systems have already been well acknowledged in computer science and neurobiology, we are reminded of the origins of the field of computer science itself, which was initiated under the direction of a cognitive scientist and a neuroscientist.
%\footnote{In the 1970s, MITÕs Artificial Intelligence Lab opened under the direction of Marvin Minskey and David Marr.}  
Fortunately, the intersections of these seemingly diverse areas of study are being specifically addressed in what some scholars are calling the field of {\em neuroaesthetics},~\cite{onians2007neuroarthistory}. For example, we know that it is the orbitofrontal and insular cortices that are involved in aesthetic judgment and that this unique feature of our executive brain functioning may distinguish us from our primate ancestors~\cite{Tsukiura}.  By examining the biological functions of the visually perceiving brain, it is possible to calculate a much more accurate understanding of the processes involved in an act of aesthetic judgment. 
\\
\indent The implications of these components of cognitive neuroscience on art and history are currently being addressed by David Freedberg in extension to his groundbreaking book on the psychological responses to art, {\em The Power of Images: Studies in the History and Theory of Response} (1989)~\cite{Freedberg89}. Other inroads on this subject from within art history have been made by Michael Baxandall through his consideration of the notion of the historically constructed period eye and his interest in the processes of visual interpretation~\cite{Baxandall1972}. The history of biological inquiries on the interpretation of art have been well summarized by John Onians in his introduction to {\em Neuroarthistory}~\cite{onians2007neuroarthistory}.
\\
\indent Interest in the psychology of seeing (in a broader sense), however, has a long history that may still be tied to Kant and the philosophical tradition. For instance, the Berlin School's theory of gestaltism that emerged in the 1890s posits that visual recognition occurs primarily on the level of whole forms as opposed to their parts. The application of gestalt psychology to art was most famously heralded by Rudolf Arnheim (1904-2007) in {\em Art and Visual Perception: A Psychology of the Creative Eye} (1954), which explored the concept of sensory knowledge through the act of seeing~\cite{Arnheim}. It is important to note that both psychologists and art historians grappling to understand the mechanisms of aesthetic interpretation have remained largely in dialogue with Kant's binary distinctions of the production knowledge. 
\\
\indent Kant's interrogations thus still underlie basic questions about machine-based intelligence: if we are able to create artificial intelligence that performs types of reasoning that we have long considered subjective, we are either more machine-like than we admit, machines have more human potential than we estimate, or these processes are, in fact, tangibly measurable and objectively determined. In essence, the debate moves to the question of determinism and free will. While most people would agree that a computer, even one that has reached automaton status and has the ability to learn from its environment, is not free, we are less willing to concede the notion of human freedom when we too are ultimately bound by our genes and environment. For eighteenth-century philosophers, reasoning, particularly in the domain of subjectivity, was tied to God through morality and universality in terms of the decisions we are perceived to freely make. These philosophies are still debated today in different terms. 
\\
\indent We would like to suggest that how we understand aesthetic judgment can still be tied to the eighteenth- and nineteenth-century philosophical tradition, yet we need to better interpret how these so-called subjective processes work, if they even are subjective, and integrate new scientific developments, such as those in neurobiology and computer science, into our conceptions of how knowledge is produced. Nonetheless, it is a paradox that developments in computer science could have pushed the humanities to reevaluate its most basic premises: for art history, it is how we determine that something is beautiful and/or important, and how objects are interrelated. Have the advances in science not provided a platform in which we can begin to understand cognition, as it is applied to aesthetics, in a radically different way than eighteenth- and nineteenth-century philosophers conceived these processes? We easily discredit the idea of humors as ruling temperaments of the body but know that Kant considered them viable and one of them as an indication of the absence of temperament~\cite{louden2002kant}.  We still read Kant for his interpretations of physical and psychological states, yet not on his theory of the phlegmatic humor.
\\
\indent Science is obviously not the only domain from which to take direction. Let us heed caution from aesthetic critics such as Julius Meier-Graefe who, in 1904, explored the problem of the dominancy of paintings in the history of art in his response to modern art and the new mediums the movement favored~\cite{Meier-Graefe}. That a machine has the ability to examine paintings does not mean that it has the capacity to understand sculpture, installation art, performance art, or land art. What would a computer make of the Christo and Jeanne-Claude installation, the {\em Wrapped Reichstag} (Figure~\ref{WrappedReichstag})? Both three- and two- dimensional computer vision programs would be able to determine the sharp edges of the building and sense its occupation of a large amount of space, either in reality, or as it appears in a photo, yet how would the significance of the wrapping of such a canonical architectural form loaded with symbolism be readily understood and quantified for qualitative analysis by a machine? 
%\begin{floatingfigure}[v]{2in}
%\end{floatingfigure}
When computer scientists one day simulate the human brain, will the machine understand the Christo and Jeanne-Claude installation? Will machine aesthetic judgment be any different than human aesthetic judgment? Who shall we give the authority to make that judgment? These are important considerations to make in our society as it adapts to the advances in artificial intelligence. Norbert Wiener's famous remarks on the effects of what he termed cybernetics remain relevant today~\cite{Wiener}. In 1950, he perceptively wrote that ``the machine, which can learn and can make decisions on the basis of its learning, will in no way be obliged to make such decisions as we should have made, or will be acceptable to us. For the man who is not aware of this, to throw the problem of his responsibility on the machine, whether it can learn or not, is to cast his responsibility to the winds~\cite{Wiener}.''

\section{Within the Limits of Probability: Computer Science and Art History Today}

This paper has considered both the limitations of computer vision research and its potential for growth in regard to its application for art history. In conclusion, we would like to underscore the current concerns that this research poses for art historians in its immediate application. We have thus highlighted three main issues that demand further attention: the use of language between fields to describe global and specific concepts, the lack of uniformity in the interpretation of art, and the separate developments within computer science and art history regarding aesthetic interpretation. 
\\
\indent Firstly, there is discomfort in the globalizing language that computer scientists use to describe their research. Rather than make claims about a computer's ability to analyze art at large, specificity as to what can be analyzed and what has been analyzed would assuage philosophical anxieties about the ontological nature of man versus the machine~\cite{Spratt2004}.\footnote{To this end, the first author presented a paper on this topic to archaeologists and art historians~\cite{Spratt2004}.}
In this paper, we have been careful to describe computer analysis of what computer scientists call visual art, a term that is not readily utilized in art history, as an analysis of paintings from some of the canonical movements in art through history in the Western tradition. Instead of framing computer vision research in broad and global terms that are unsupportable (from the humanities' perspective), demonstrating the potential of this technology through specific examples allows art historians to consider its value in ways that don't interfere with their critical approach of analysis. If we can shift the onus of interpretation to the art historians, computer scientists would likely find art historians more willing to embrace computer vision technology.
\\
\indent While computer vision research has been instrumental in art conservation applications, it has not been utilized by art historians for more aesthetically based interpretations. 
Not surprisingly, our surveys further confirmed the apprehension in art history to the developments that computer vision offers in the realm of subjective interpretation. 
We would therefore like to propose that computer scientists collaborate with art historians on specific projects. Research that concerns the analysis of a multitude of images related to one artist or movement could be facilitated by the current capabilities of computer vision technology. The ability to compute perspective coherence, lighting and shading strategies, brushstrokes styles, and semantic points of similarity could, for example, aid the analysis of a large group of Italian drawings with unclear authorship. Similarly, the application of this technology for the identification of icon workshops that utilized the same iconographic templates in the context of Medieval, Byzantine, or Post-Byzantine devotional images would be extremely useful if a large data set of icons from diverse collections that are not readily accessible to the public could be brought together.
 Recent collaborations of this nature have already been initiated and should continue~\cite{Jia12,hughes2012empirical}. While this type of collaboration lies in the domain of connoisseurship more than what one would term art history, it seems clear that working within the realm of current capabilities in computer vision technology is the best way to build a collaboration between the fields that would eventually ignite a more philosophical understanding of these methods and their bearing on aesthetic theory.  
 \\
\indent The second issue regarding the immediate application of computer vision research in the domain of aesthetics concerns the way the social history of an object and the emotional engagement to art is calculated. In art history, the degree to which the context in which a work of art is produced should matter. How can a computer quantify the social history of a painting or the material means of its production? It is exactly this point that the critical theorists of art raised more than a century ago regarding the nature of art ``both context-bound and yet also irreducible to its contextual conditions~\cite{podro1984critical}.''  To quote the art historian Michael Podro,  ``Either the context-bound quality or the irreducibility of art may be elevated at the expense of the other. If a writer diminishes the sense of context in his concern for the irreducibility or autonomy of art, he moves towards formalism. If he diminishes the sense of irreducibility in order to keep a firm hand on extra-artistic facts, he runs the risk of treating art as if it were the trace or symptom of those other facts~\cite{podro1984critical}.'' If art is treated autonomously, as having an independent progression in the realm of form, its history is purely stylistic. For the critical theorists, this extreme was considered an aesthetic failure, as judgment requires morality and thus is tied to value-based interpretations of art on the level of object analysis~\cite{podro1984critical}.  
\\
\indent Furthermore, if our understanding of the history of art is related to the emotional response that an object elicits, how can a computer mimic human affect? On the other hand, the developments in computer vision technology and neurobiology suggest a new understanding of the very mechanisms of emotion. That what we have understood as subjective processes may in fact be objectively determined problematizes the argument that computers can never achieve the capacities beholden to the contemplative human mind.
These issues have been addressed in the recent surge of philosophical research on creativity~\cite{Boden2014,Bohm96}.
%\footnote{See, Elliot Samuel Paul and Scott Barry Kaufman Eds., The Philosophy of Creativity, New Essays, (Oxford: Oxford University Press, 2014). See in particular, the chapter, ÒCreativity and Artificial Intelligence,Ó especially the section, ÒWhat Creativity-Denying Features do Computers Possess,Ó pp. 229-232~\cite{Boden2014}. While the physicist and theorist David Bohm (1917-1992) was a pioneer in the philosophy of creativity, his discourse on the relationship of science and art addressed the role of an artist's creative capacities, not the ability of an art historian to interpret objects or the nature of the humanities in relation to the hard sciences. See, David Bohm, On Creativity, (New York: Routledge, 1996).} 
 \\
\indent In essence, there is no singular correct interpretation of a work of art within art history, as multiple theories and methodologies place differing emphases on style, content, and context. To date, computer vision research offers predominantly stylistic interpretations of paintings that only recently have begun to include iconographic considerations. While these tools have allowed us to categorize paintings into broad genres and chronologies, computer science is currently unable to offer more immediate associations regarding the specific social history of an object and the degree to which these conditions influenced the final product. In the same vein, certain periods or genres are more amenable to some theoretical approaches than to others. For example, abstract expressionism, which is highly concerned with the role of form over content, naturally accommodates the high degree of stylistic interpretation that computer vision offers. Within modern art, computer vision research might have the potential to offer unexpected insights on the level of style.  
\\
\indent Due to the use of broad data sets, it is not surprising that computer scientists have noticed some far-reaching stylistic influences. For instance, automatic influence detection demonstrated the ability to detect less overt connections between artists such as Eugene Delacroix's not-so-widely-known influence from El Greco both in terms of color and expressiveness~\cite{Abe2013MM4CH}. While this observation highlights the remarkable subtleties of interpretation that computer vision is capable of generating, this type of analysis is of less use to an art historian than a more specific study, such as what an analysis of Kazimir Malevich's fairly uniform appearing Suprematist paintings might reveal in regard to style.
\\
\indent The last critical issue that emerges concerns the way we locate and attribute the onus of interpretation in computer vision analysis. To what degree can we ascribe the detection of influence or artistic merit to a machine when it was the computer scientists that wrote the programming that associated certain visual components with particular markers of identity? At what point in the process of training the program to make its own judgments does the machine develop autonomy, if ever? If computer scientists can be charged with owning the responsibility of artistic interpretation at the level of programming input, why wouldn't art historians be involved at this level of the research? While there is no question that at this stage of development within computer science that programs have demonstrated the ability to take on an autonomous quality based on what they have been taught, are these innovations so advanced at this point in time that we can consider them on par to human judgment? Unfortunately,  aesthetic interpretation in computer science is developing in isolation from the aesthetic discourse in philosophy and art history. If the humanities were able to more clearly understand the use-value of computer vision research and art historians were able to collaborate with computer scientists as machine-based aesthetic interpretation develops, both fields would benefit. 
\\
\indent That a computer is able to measure art aesthetically challenges the field of art history to reexamine its own aesthetic constructs. David Hume pontificated that ``beauty is no quality in things themselves: it exists merely in the mind which contemplates them; and each mind perceives a different beauty~\cite{hume1757standard}.''  If the interpretation of art lies in the eyes of the beholder and is thus a subjectively determined process that is associated with feeling, how can we understand the development of autonomous aesthetic evaluation from a computer without reevaluating the processes of human aesthetic judgment and emotion? Awareness of these concepts could equally steer the direction of computer vision in terms of its abilities to provide immediate practical applications to the field of art history rather than taking on the uncomfortable guise of a virtual art historian. Our survey confirmed that both computer scientists and art historians agree that the humanities should be more digitized; however, before art historians are willing to believe that it is possible to analyze art with a computer in terms of beauty, style, dating, and relative influence in the development of art through history, we must revisit the concept of aesthetic judgment.

\clearpage

\bibliographystyle{splncs}
\bibliography{visualart}

%\bibliographystyle{splncs03}
%\bibliography{egbib}
\end{document}